\pdfoutput=1
\documentclass[10pt]{article}
\usepackage[preprint]{tmlr}
\usepackage{amsmath,amssymb,amsfonts}
\usepackage{graphicx,booktabs,caption,hyperref,url}
\usepackage{array}
\usepackage[T1]{fontenc}

\newcommand{\lamwd}{\lambda_{\mathrm{wd}}}

\title{Weibull Weight-Scale Parameter Evolution under AdamW Training Dynamics}

\author{\name Tiexin Ding\thanks{Independent Researcher. Email: \texttt{tiexinding@gmail.com}}}

\begin{document}
\maketitle

\begin{abstract}
Building on a two-parameter Weibull framework for diagnosing transformer weight distributions, we study why the Weibull weight-scale parameter $\lambda$ grows, overshoots, and then relaxes during AdamW training. We derive a \textbf{leading-order} three-force decomposition of the squared weight norm from the AdamW update: an alignment force measuring the correlation between weights and the adaptive update direction, an injection force from adaptive step magnitude, and a decay force from decoupled weight decay. On self-trained Pythia-70M models with ground-truth optimizer moments, alignment dominates the rise phase, contributing 88--94\% of the absolute force budget across four random seeds and remaining robust to super-weight removal. Near saturation, alignment and decay approach balance, explaining the transition from weight-scale growth to relaxation. These force dynamics directly govern the squared-norm component underlying $\lambda(t)$; the remaining RMS-to-Weibull reconstruction offset is measurable and decomposes into bridge and integration components, totaling approximately 5--6\% in densely sampled regions. To extend the analysis to real models where optimizer moments are unavailable, we introduce a spline displacement method that recovers the alignment force from sparse checkpoints with approximately 92--94\% accuracy, about twice the naive two-point baseline. We further observe that the peak value of $\lambda(t)$ varies with training-data coherence in our experiments, suggesting a data-dependent component of weight-scale growth that we leave to a controlled follow-up study. Code and data are available at \url{https://github.com/tiexinding/NPM-Weibull-public}.
\end{abstract}

\section{Introduction}
Understanding how neural network weights evolve during training is central to understanding optimization, generalization, and architecture design. A previous two-parameter Weibull framework introduced a distribution-level diagnostic for transformer weight magnitudes: the shape parameter $k$ describes the body shape of the element-wise weight distribution, while the scale parameter $\lambda$ describes its weight-scale evolution \citep{npmweibull2026paper1}. Across 12 model entries spanning 7 architectural families, that framework identified two regularities: (1) for Transmission-class components, $k$ remains approximately locked near its initialization anchor $k_0 \approx 1.20$, and (2) $\lambda$ follows a characteristic rise, overshoot, relaxation, and floor trajectory during training \citep{npmweibull2026paper1}.

These observations raise a natural follow-up question: \emph{why does the Weibull scale parameter $\lambda$ rise and then relax?}

This question sits between two levels of description. AdamW dynamics are defined at the level of individual parameters and optimizer states \citep{loshchilov2019decoupled}, while the Weibull parameters $k$ and $\lambda$ are distribution-level summaries of entire weight matrices \citep{npmweibull2026paper1}. The previous Weibull framework established $\lambda$ as a stable weight-scale coordinate, but it did not identify which optimizer-level forces move the model along that coordinate. Conversely, prior work on AdamW and weight-decay dynamics has analyzed norm-level quantities such as weight-update alignment, adaptive step magnitude, rotational equilibrium, and weight decay \citep{kosson2024rotational, chou2025correction, kosson2025weight}. These quantities live at the squared-norm level. They do not by themselves specify how a higher-level Weibull statistic inherits the observed rise, overshoot, relaxation, and floor trajectory. Bridging these two levels is the goal of this paper.

This paper builds that bridge. We derive a leading-order three-force decomposition of AdamW squared-norm dynamics from the decoupled AdamW update \citep{loshchilov2019decoupled}: an alignment force determined by the correlation between weights and the adaptive update direction, an injection force from adaptive step magnitude, and a decay force from decoupled weight decay. We measure these forces directly on self-trained transformers with ground-truth optimizer moments. During the rise phase, alignment dominates the absolute force budget; near saturation, alignment and decay approach balance. This alignment-to-decay transition explains the dominant weight-scale component underlying the observed $\lambda(t)$ trajectory.

The connection to Weibull $\lambda$ requires an explicit bridge. The force decomposition directly governs the squared weight norm, not the fitted Weibull scale parameter. The $k$-lock established by the Weibull framework makes this bridge possible: when $k$ remains approximately constant, RMS and $\lambda$ differ by an approximately fixed multiplicative factor, $\sqrt{\Gamma(1 + 2/k)}$. We therefore quantify the remaining RMS-to-Weibull reconstruction offset rather than treating the two quantities as identical. This also connects the transient $\lambda$ trajectory to prior observations that terminal weight scale is shaped by learning rate and weight decay \citep{fan2025robust}.

We also address the practical setting in which optimizer moments are unavailable. Public model releases usually provide sparse weight checkpoints, not AdamW first and second moments. To recover force information from such checkpoints, we introduce a spline displacement method based on the AdamW displacement identity. On self-trained models where ground-truth optimizer moments are available for validation, the method recovers the alignment force with approximately 92--94\% accuracy, about twice the naive two-point baseline.

Finally, we report an exploratory observation that the absolute peak of $\lambda(t)$ varies with training-data coherence. This observation is consistent with broader evidence that data diversity can shape weight-space structure \citep{ba2024data}, but our setting does not yet isolate the data-diversity axis with controlled mixtures. We therefore treat this as suggestive evidence for a data-dependent component of weight-scale growth and leave controlled data-mixture experiments to future work.

\begin{figure}[t]\centering
  \includegraphics[width=\linewidth]{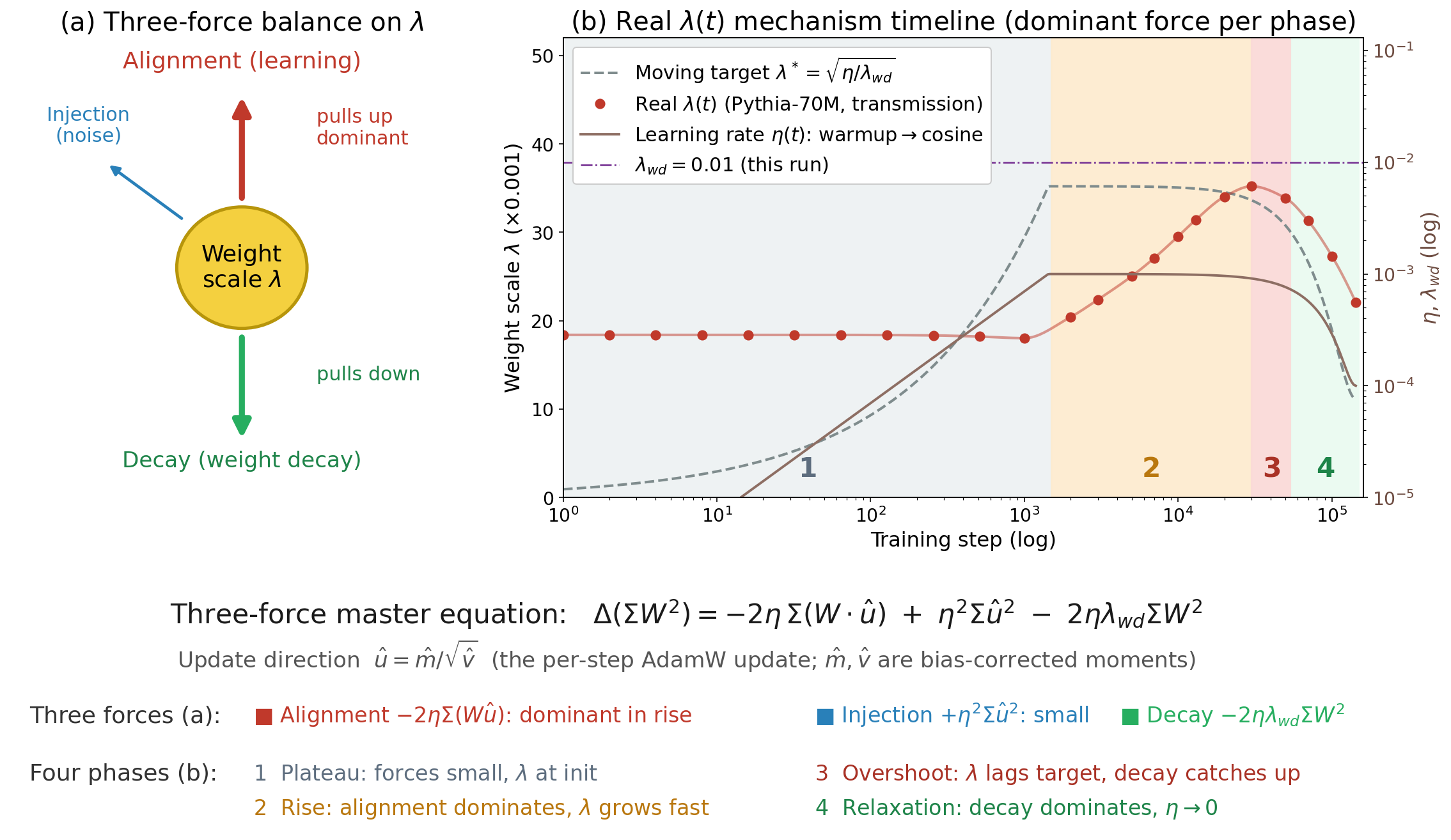}
  \caption{Mechanism overview. (a) The three forces acting on the squared weight scale and, through the $k$-lock bridge, on the Weibull scale $\lambda$: alignment (the learning signal, dominant 88--94\% in the rise phase) pulls up, decay pulls down, injection is a small noise term. (b) The resulting $\lambda(t)$ trajectory on real Pythia-70M over four phases, against the moving target $\lambda^*\propto\sqrt{\eta/\lambda_{\mathrm{wd}}}$ (the learning rate $\eta$ peaks at step 1430 and then cosine-decays; $\lambda_{\mathrm{wd}}=0.01$ is the value used in this run). The 88--94\% rise-phase alignment share is the measured value for this setting. Master equation derived in Section~3.}
  \label{fig:hero}
\end{figure}

Figure~\ref{fig:hero} gives a schematic overview of the three-force mechanism and the resulting $\lambda(t)$ trajectory.

\textbf{Contributions:}
\begin{enumerate}
  \item \textbf{Weibull-to-AdamW connection.} We connect the Weibull weight-scale parameter $\lambda$ to AdamW squared-norm dynamics, showing that $\lambda(t)$ tracks the weight-scale component of a leading-order force budget up to a measurable RMS-to-Weibull reconstruction offset.
  \item \textbf{Rise-phase force budget.} Using self-trained Pythia-70M models with ground-truth optimizer moments, we show that alignment dominates the rise phase, contributing 88--94\% of the absolute force budget across four random seeds and remaining robust to super-weight removal.
  \item \textbf{Alignment-to-decay transition.} We show that the transition from $\lambda$ growth to relaxation corresponds to a force-balance transition: alignment dominates early, while alignment and decoupled weight decay approach balance near saturation.
  \item \textbf{Sparse-checkpoint force recovery.} For real models without optimizer moments, we introduce a spline displacement method that approximately recovers the alignment force from sparse checkpoints, achieving 92--94\% recovery and about 2$\times$ the naive two-point baseline.
\end{enumerate}

\section{Setup}
\subsection{AdamW}
We train with AdamW \citep{loshchilov2019decoupled}. The per-parameter update is:
\begin{equation}
W_{t+1} = W_t - \eta_t \cdot \hat{u}_t - \eta_t \cdot \lamwd \cdot W_t
\end{equation}
where $\hat{u}_t \equiv \hat{m}_t / (\sqrt{\hat{v}_t} + \varepsilon)$ is the effective adaptive update direction, $\hat{m}_t = m_t/(1-\beta_1^t)$ is the bias-corrected first moment, $\hat{v}_t = v_t/(1-\beta_2^t)$ is the bias-corrected second moment, and $\lamwd$ is the weight decay coefficient.

\subsection{The Weibull Scale Parameter $\lambda$}
Following \citet{npmweibull2026paper1}, we fit each weight matrix's element-wise magnitude distribution $|W|$ to a two-parameter Weibull$(k, \lambda)$ by least squares on the Weibull probability plot, using its middle-80\% trim protocol.

We track the fitted scale parameter $\lambda$, \emph{not} the raw weight RMS $\sigma = \sqrt{\Sigma W^2/N}$. The two are related by $\sigma = \lambda\sqrt{\Gamma(1+2/k)}$. Because $k$ stays locked near $k_0 \approx 1.20$ \citep[confirmed here in \S6]{npmweibull2026paper1}, the factor $\sqrt{\Gamma(1+2/k)}$ is nearly constant ($\approx 1.23$); thus $\lambda$ and $\sigma$ follow the same trajectory up to an approximately fixed multiplicative factor, with $\sigma$ about 23\% larger than $\lambda$ when $k \approx 1.20$. We report $\lambda$ via the Weibull fit throughout, and quantify the residual between the RMS-implied $\lambda$ and the fitted $\lambda$ in Appendix~B.

\emph{Notation.} Here $\sigma = \sqrt{\Sigma W^2/N}$ is the per-layer weight RMS, distinct from the ratio metric $\sigma_{\mathrm{ratio}} = \sigma_{\mathrm{in}}/\sigma_{\mathrm{out}}$ of \citet{npmweibull2026paper1}.

\subsection{Dynamics Are Derived for $\Sigma W^2$}
The three-force decomposition (\S3) acts on the raw second moment $\Sigma W^2 = N\sigma^2$, the quantity that arises naturally from squaring the AdamW update. The $k$-lock bridges this to the Weibull scale: because $k$ is approximately constant, force dynamics on $\sigma^2$ determine the dominant RMS component underlying Weibull $\lambda(t)$, up to the RMS-to-Weibull reconstruction offset quantified in Appendix~B.

\subsection{Data}
This paper uses three types of models, summarized in Table~\ref{tab:models}.

\begin{table}[t]\centering\footnotesize\setlength{\tabcolsep}{5pt}
\begin{tabular}{>{\raggedright\arraybackslash}p{2.6cm}>{\raggedright\arraybackslash}p{2.0cm}>{\raggedright\arraybackslash}p{2.0cm}>{\raggedright\arraybackslash}p{4.5cm}}
\toprule
Source & Model & Nature & Role in this paper \\
\midrule
Real Pythia checkpoints (EleutherAI) & 70m, 160m, 410m, 1B & Real published models & Phenomenology baseline: $\lambda(t)$ trajectories (overshoot$\to$relax$\to$floor) \\
Self-trained Pythia-arch (V1b) & 70m, random-init & Controlled ground truth & Primary evidence: three-force decomposition, force budget, spline method, $k$-lock \\
Self-trained Llama-arch & 70m, random-init & Controlled architecture transfer & Cross-architecture robustness test \\
\bottomrule
\end{tabular}
\caption{Three model types used in this paper.}
\label{tab:models}
\end{table}

\textbf{Self-trained Pythia-70M (V1b).} We train a GPT-NeoX model with the Pythia-70M configuration from random initialization on wikitext-103. Training: lr~$10^{-3}$, $\lamwd=0.01$, $\beta=(0.9, 0.999)$, warmup 200 steps then cosine decay to $0.1\times$, batch 24, sequence length 512, 20{,}000 steps, checkpoints every 250 steps. This provides ground-truth $\hat{m}$ and $\hat{v}$.

\textbf{Real Pythia ($\lambda(t)$ baseline).} We use the fitted Weibull $\lambda(t)$ trajectories for Pythia-70M/160M/410M/1B reported by \citet{npmweibull2026paper1}.

\textbf{Self-trained Llama-style model.} We train a Llama-style model (SwiGLU, RMSNorm, full RoPE, no bias) from random initialization on the same wikitext-103 corpus.

\textbf{Availability.} The Weibull fitting library, the \citet{npmweibull2026paper1} database of fitted trajectories, and the code for the force decomposition and spline recovery are released with the companion NPM-Weibull repository: \url{https://github.com/tiexinding/NPM-Weibull-public}. The database is also available at \url{https://huggingface.co/datasets/TiexinDing/NPM-Weibull-DATABASE-v9_1}.

\section{Three-Force Decomposition}
\subsection{Derivation}
Starting from the AdamW update $W_{t+1} = W_t - \eta \cdot \hat{u} - \eta \cdot \lamwd \cdot W_t$:
\begin{equation}
(w_{t+1})^2 - w_t^2 = \eta^2\hat{u}^2 - 2\eta \cdot \hat{u} \cdot w - 2\eta \cdot \lamwd \cdot w^2 + O(\eta^2 \cdot \lamwd)
\end{equation}
The dropped higher-order terms are quantified in Appendix~A and contribute less than $0.001\%$ of the absolute force budget in our runs. Summing over all $N$ parameters:
\begin{equation}
\Delta(\Sigma W^2) = -2\eta\langle W, \hat{u}\rangle + \eta^2\|\hat{u}\|^2 - 2\eta \cdot \lamwd \cdot \Sigma W^2
\end{equation}

We decompose into three forces:
\begin{itemize}
  \item \textbf{Alignment force} $F_{\mathrm{align}} = -2\eta\langle W, \hat{u}\rangle$ --- positive when $\langle W, \hat{u}\rangle < 0$, driving net weight growth.
  \item \textbf{Injection force} $F_{\mathrm{inj}} = +\eta^2\|\hat{u}\|^2$ --- always positive, proportional to the squared adaptive step magnitude.
  \item \textbf{Decay force} $F_{\mathrm{decay}} = -2\eta \cdot \lamwd \cdot \Sigma W^2$ --- always negative, proportional to the current weight scale.
\end{itemize}

\subsection{Physical Meaning}
\textbf{Alignment force} measures whether the AdamW adaptive step is correlated with the current weight direction. In the rise phase, $\langle W, \hat{u}\rangle < 0$ --- equivalently $\langle W, \Delta W\rangle > 0$, since $\Delta W = -\eta\hat{u}$ --- meaning the update points \emph{along} the current weight direction.

\textbf{Injection force} is the squared magnitude of the adaptive step, representing stochastic gradient noise.

\textbf{Decay force} is the structural contribution from decoupled weight decay, proportional to the current weight scale.

\section{Force Budget}
\subsection{Rise Phase: Alignment Dominates}
On self-trained Pythia-70M, we compute all three force contributions at each training step using ground-truth $\hat{m}$ and $\hat{v}$. During the rise phase (steps $\leq$ 5,000), the alignment force contributes \textbf{88--94\% of the absolute force budget}. This result is robust across four random seeds and to removal of super-weight outliers (identified as the top 0.1\% by $|W\cdot\hat{u}|$ magnitude; removal changes alignment share by only 0.4 percentage points, from 93.8\% to 93.4\%). The injection force contributes only $\sim$4\%, and the decay force $\sim$2--7\%.

\begin{figure}[t]\centering
  \includegraphics[width=\linewidth]{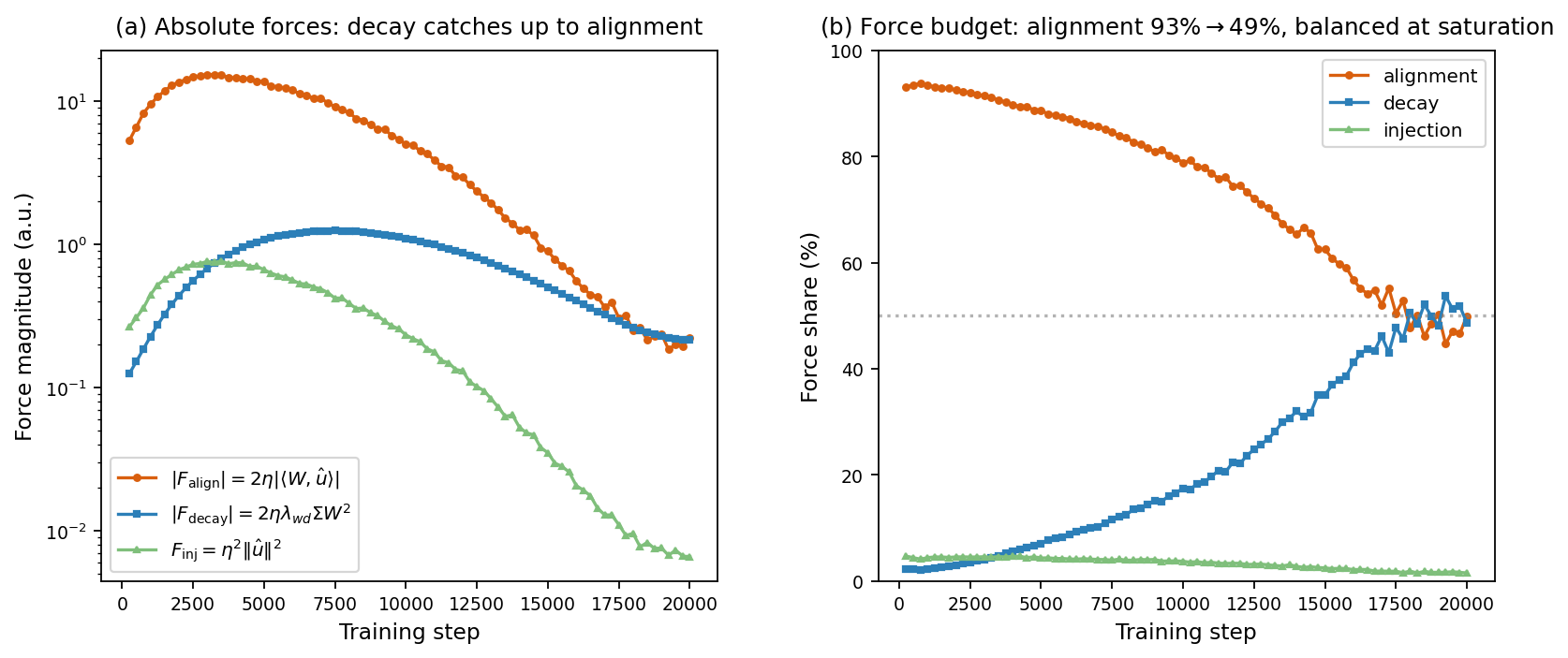}
  \caption{Three-force budget on self-trained Pythia-70M. (a) Absolute force magnitudes (log scale): alignment dominates the rise, decay catches up at saturation. (b) Fractional shares: alignment falls $93\%\to49\%$ while decay rises $2\%\to49\%$, balancing (net $\approx0$) at saturation.}
  \label{fig:three-force-budget}
\end{figure}

Figure~\ref{fig:three-force-budget} shows the three-force budget and fractional composition during training.

\begin{table}[t]\centering\footnotesize
\begin{tabular}{rccccl}
\toprule
step & $\sigma$ & align\% & inj\% & decay\% & phase \\
\midrule
250 & 0.021 & 93.1 & 4.7 & 2.2 & early rise \\
1{,}250 & 0.031 & 93.1 & 4.5 & 2.4 & rise \\
2{,}500 & 0.045 & 92.1 & 4.5 & 3.4 & rise \\
5{,}000 & 0.066 & 88.7 & 4.3 & 7.0 & mid-rise \\
10{,}000 & 0.084 & 78.9 & 3.7 & 17.5 & late/peak \\
20{,}000 & 0.087 & 49.9 & 1.5 & 48.6 & saturation \\
\bottomrule
\end{tabular}
\caption{Full force-budget trajectory on self-trained Pythia-70M. Alignment falls $93\%\to49\%$ while decay rises $2\%\to49\%$, balancing (net $\approx0$) at saturation; injection stays $\sim$4\%.}
\label{tab:trajectory}
\end{table}

\textbf{Cross-architecture robustness.} We repeat the controlled self-training with a Llama-style architecture (SwiGLU/RMSNorm/RoPE/no bias), holding data, optimizer, and scale fixed. The alignment force again dominates the rise phase: \textbf{88.0--93.4\%} for steps $\leq$ 5,000, nearly identical to the Pythia-style result (88.7--93.8\%).

\begin{figure}[t]\centering
  \includegraphics[width=\linewidth]{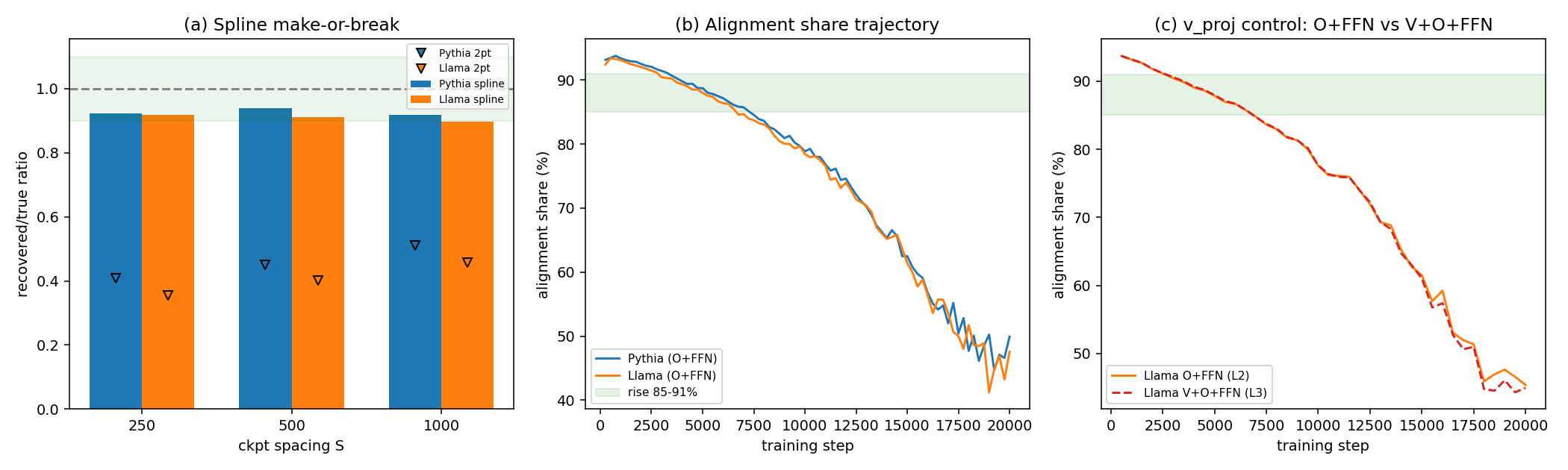}
  \caption{Cross-architecture robustness. Rise-phase alignment share is $88$--$94\%$ (cross-seed) for both Pythia-style and Llama-style, tracking within $\sim$1\% at every step.}
  \label{fig:cross-arch}
\end{figure}

Figure~\ref{fig:cross-arch} compares the rise-phase alignment share across the two architectures.

\begin{figure}[t]\centering
  \includegraphics[width=\linewidth]{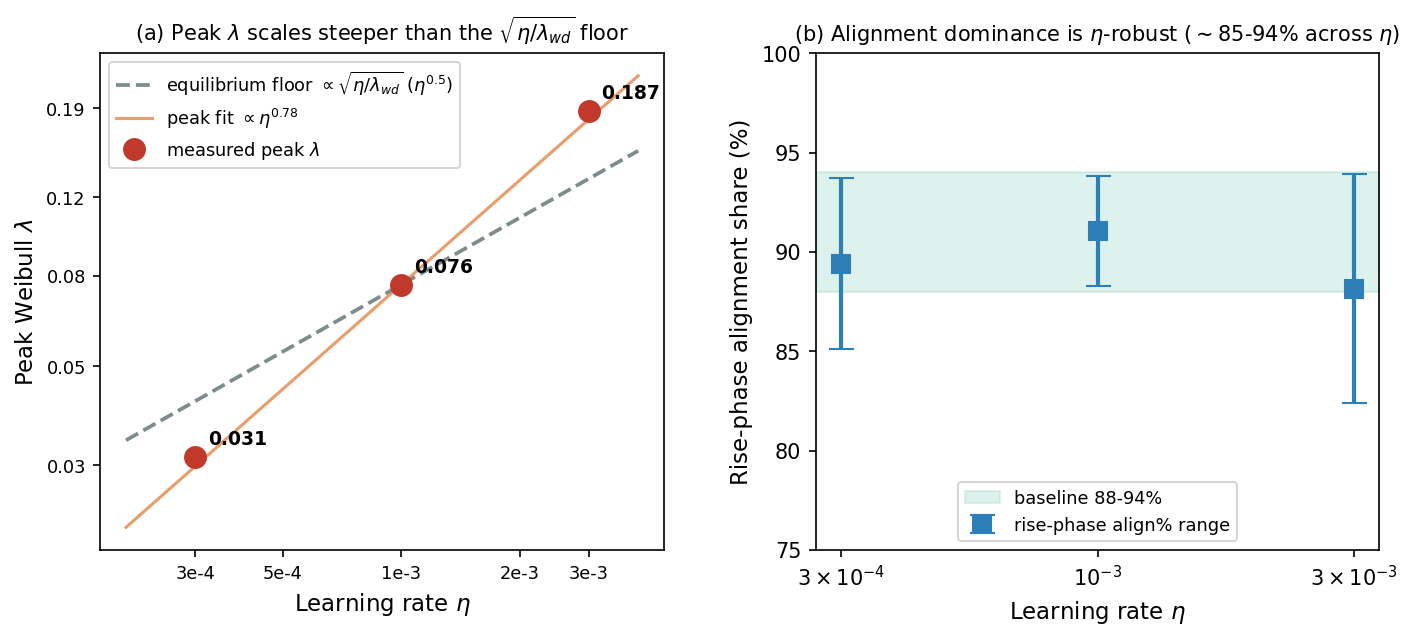}
  \caption{Learning-rate sweep on self-trained Pythia-70M. (a) Peak Weibull $\lambda$ vs $\eta$ (log-log): measured peaks (0.031, 0.076, 0.187) approximately follow $\eta^{0.78}$ (three points, not a fitted power law), steeper than the equilibrium-floor scaling $\sqrt{\eta/\lambda_{\mathrm{wd}}}\propto\eta^{0.5}$; at $\eta=3\times10^{-3}$ the peak exceeds the floor, confirming the peak is a transient overshoot above the steady state. (b) The rise-phase alignment-share center points for all three $\eta$ fall within the 88--94\% baseline band (error bars show cross-seed spread), confirming the alignment-dominance mechanism is learning-rate-robust.}
  \label{fig:eta-sweep}
\end{figure}

\textbf{Learning-rate robustness.} We additionally sweep the learning rate over three values (Figure~\ref{fig:eta-sweep}). The peak $\lambda$ increases with $\eta$, while the rise-phase alignment-share center points for all three $\eta$ fall within the 88--94\% baseline band, indicating that alignment dominance is not specific to a single learning-rate setting. The observed peak scaling is steeper than the equilibrium-floor scaling, consistent with the peak being a transient overshoot rather than the terminal steady state.

\begin{figure}[t]\centering
  \includegraphics[width=0.78\linewidth]{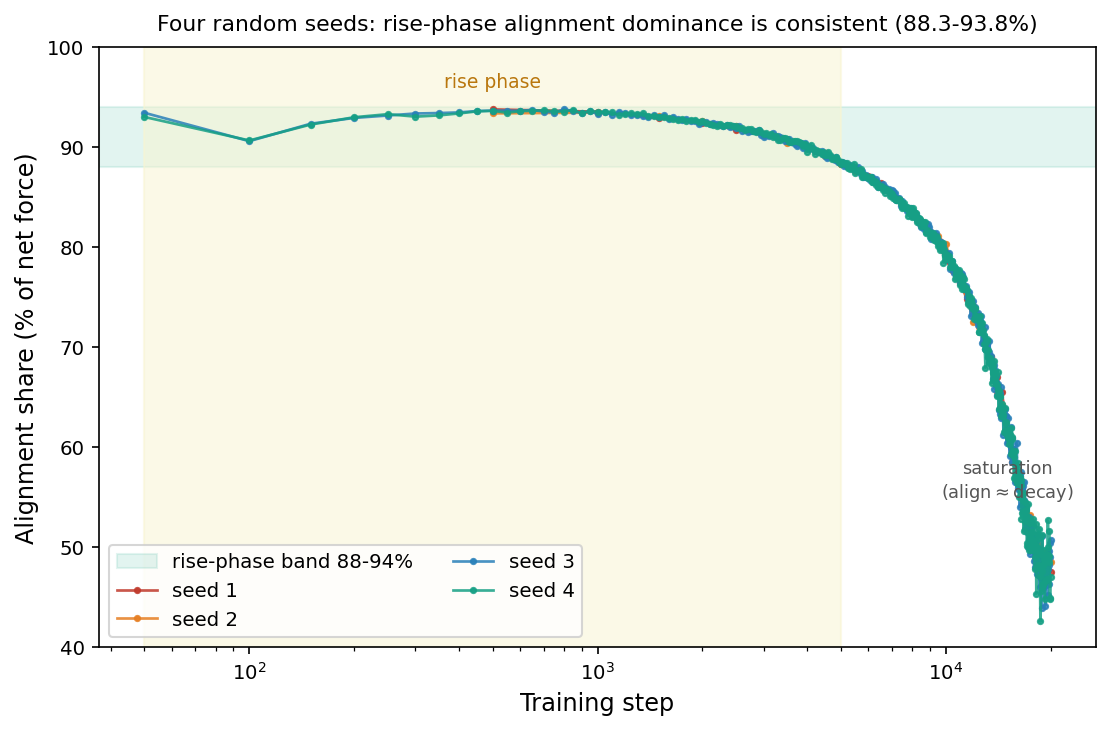}
  \caption{Alignment share over training for four random seeds. The rise-phase (steps $\leq$5,000) alignment dominance is consistent at 88.3--93.8\% across all seeds (curves nearly indistinguishable); all four then decay to the align$\approx$decay saturation balance, confirming the force-budget result is seed-robust.}
  \label{fig:4seed}
\end{figure}

\textbf{Seed robustness.} Figure~\ref{fig:4seed} shows the full alignment-share trajectory for all four random seeds. The curves nearly overlap during the rise phase, with alignment contributing 88.3--93.8\% of the absolute force budget before step 5{,}000, and all seeds move toward the same late-stage align--decay balance.

\subsection{Full Training Trajectory: Force Dynamics Explain the Weight-Scale Component of $\lambda(t)$}
The force dynamics explain the dominant squared-norm/RMS component underlying the overshoot$\to$relaxation$\to$floor pattern of $\lambda(t)$:
\begin{itemize}
  \item \textbf{Rise}: $F_{\mathrm{align}} > |F_{\mathrm{decay}}|$ $\to$ positive net force $\to$ $\Sigma W^2$ grows, and $\lambda$ follows through the $k$-lock bridge.
  \item \textbf{Peak}: $F_{\mathrm{align}} \approx |F_{\mathrm{decay}}|$ $\to$ net force $\approx$ 0 $\to$ weight-scale growth stalls.
  \item \textbf{Relaxation}: $F_{\mathrm{align}} < |F_{\mathrm{decay}}|$ $\to$ negative net force $\to$ $\Sigma W^2$ shrinks, and $\lambda$ follows up to the RMS-to-Weibull reconstruction offset.
  \item \textbf{Floor}: the force budget approaches a low-drift balance that maintains the post-training weight scale.
\end{itemize}

Figure~\ref{fig:three-force-budget} shows the fractional force composition over time.

\begin{figure}[t]\centering
  \includegraphics[width=\linewidth]{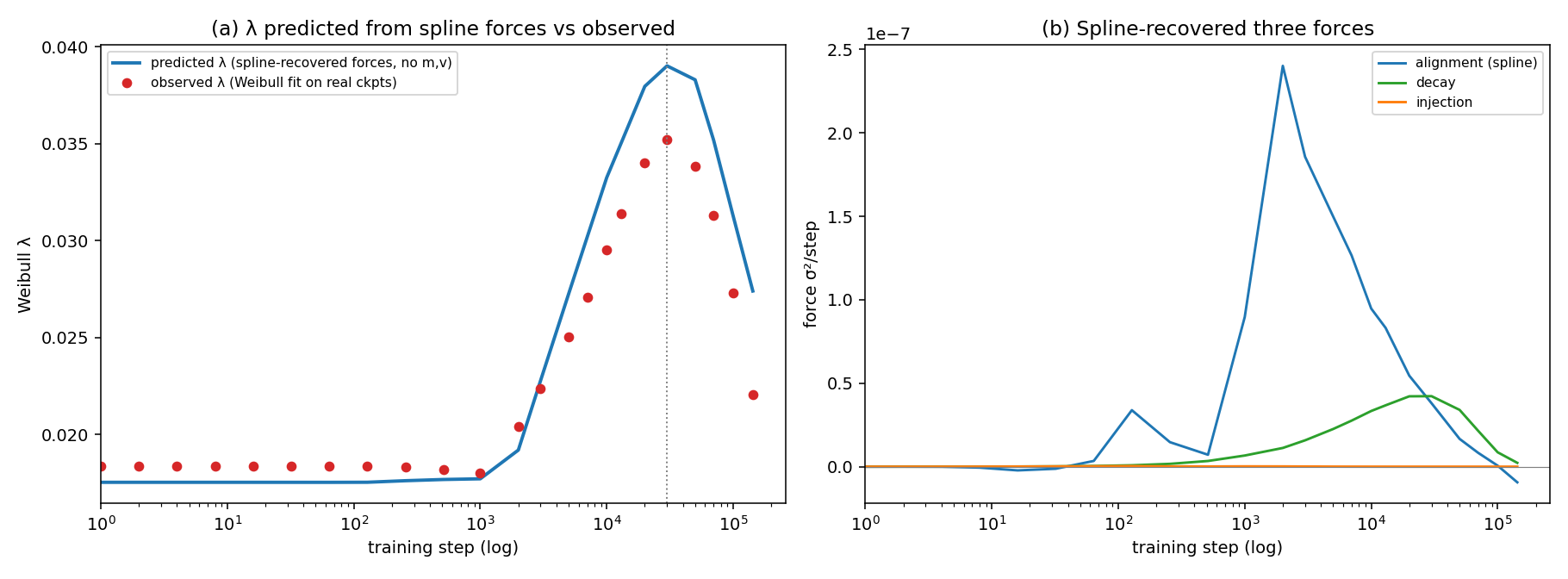}
  \caption{Real Pythia-70M application. Spline-recovered $\lambda(t)$ closure reproducing the published trajectory (aggregate peak $\lambda\approx0.035$ at step 30k). Per-block $k=1.190$ (in-band); aggregate $k=1.162$ is a pooling artifact.}
  \label{fig:realpythia}
\end{figure}

\textbf{Real-checkpoint closure.} Figure~\ref{fig:realpythia} applies the validated sparse-checkpoint recovery pipeline to the published Pythia-70M trajectory. Although optimizer moments are unavailable for this real model, the spline-recovered force trajectory reproduces the observed $\lambda(t)$ closure, including the peak near step 30k and the subsequent relaxation. This should be interpreted as checkpoint-level consistency evidence, while recovery accuracy itself is validated on self-trained runs with optimizer moments.

\begin{figure}[t]\centering
  \includegraphics[width=\linewidth]{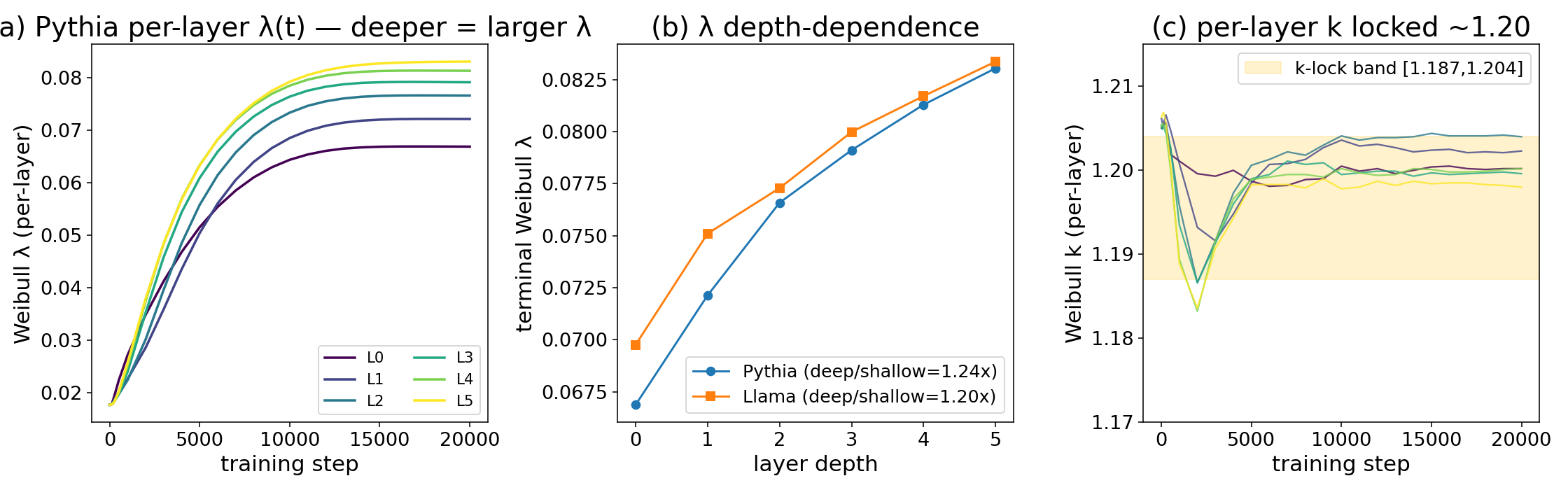}
  \caption{Per-layer $\lambda(t)$ and depth structure (self-trained Pythia and Llama-style). Per-layer $k$ stays in $[1.178,1.216]$ (median $1.20$); deep layers reach larger terminal $\lambda$ ($1.14$--$1.19\times$ shallow).}
  \label{fig:per-layer}
\end{figure}

\begin{figure}[t]\centering
  \includegraphics[width=\linewidth]{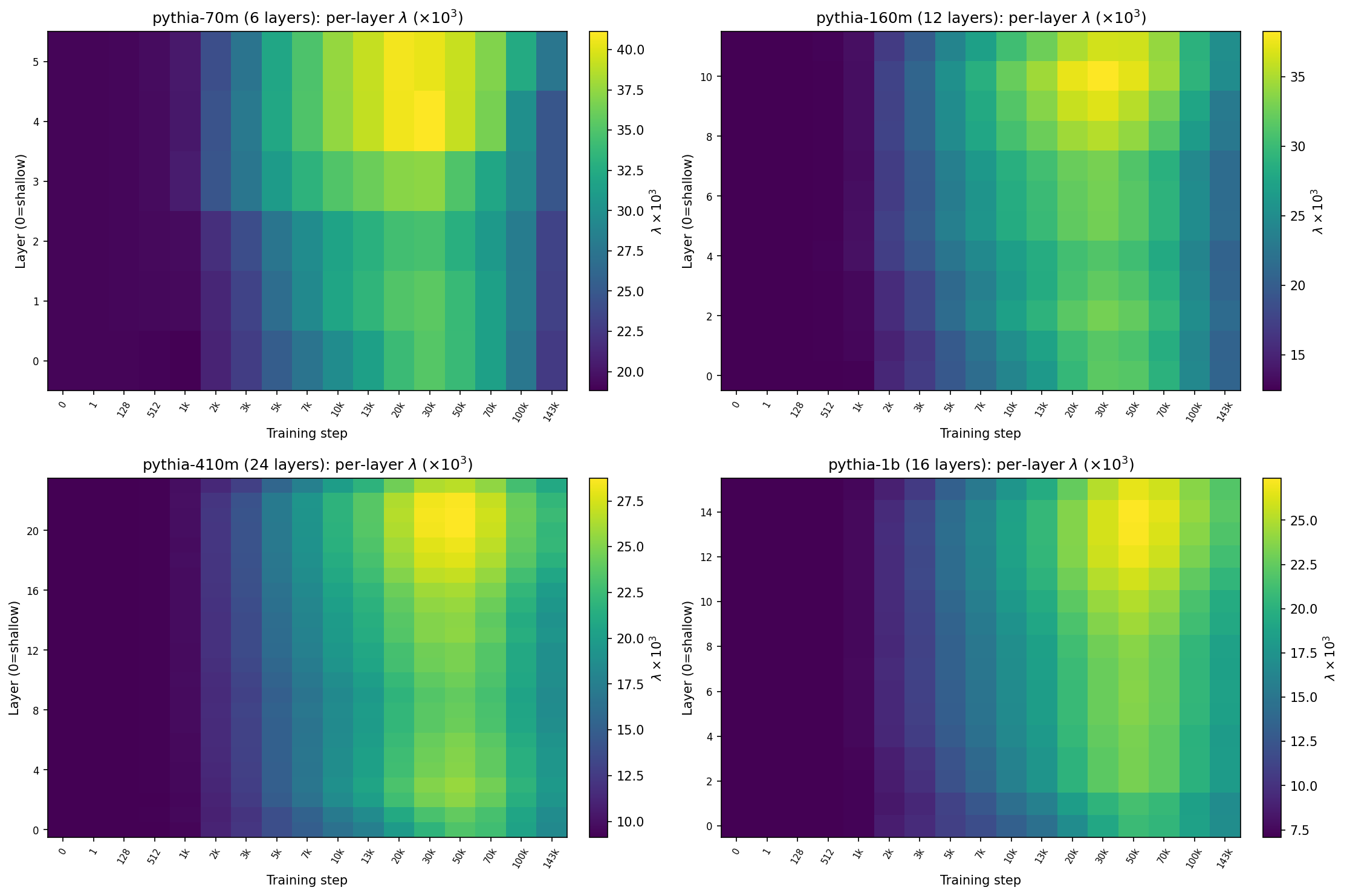}
  \caption{Per-layer $\lambda(t)$ across four real Pythia sizes (70m/160m/410m/1b). Two patterns hold across sizes: overshoot in time (peak $\sim$20k--50k then relax) and deeper-layers-higher ($1.14$--$1.27\times$).}
  \label{fig:heatmap}
\end{figure}

\textbf{Layer and size structure.} Figures~\ref{fig:per-layer} and~\ref{fig:heatmap} show that the $\lambda(t)$ trajectory is not only an aggregate effect. In self-trained Pythia-style and Llama-style models, the rise-and-relaxation pattern appears at the layer level, with deeper layers reaching larger terminal $\lambda$ values. The same qualitative pattern appears across published Pythia sizes from 70M to 1B, where Transmission-class layers peak around 20k--50k steps and then relax toward a floor. These per-layer and cross-size trajectories provide the real-model phenomenology that the self-trained force-budget analysis explains at the squared-norm level.

\section{Measurement and Recovery Methods}
\subsection{Ground Truth from Self-Training}
Self-training provides ground-truth $\hat{m}$ and $\hat{v}$ at every step, enabling direct computation of all three forces. We verify the decomposition's self-consistency by closed-loop integration: starting from the initial squared norm, integrating the leading-order squared-norm recurrence forward, and mapping the resulting RMS trajectory to $\lambda$ through the Weibull bridge, we compare the predicted $\lambda(t)$ to the observed trajectory. On the self-trained transformer V1b, logRMSE = 0.053 (Pythia) and 0.056 (Llama). Across both architectures the three-force decomposition reproduces the fitted Weibull-$\lambda$ trajectory with a consistent systematic offset of $\sim$4--6\% (logRMSE $\approx$ 0.05; logRMSE uses the natural log, so for small relative errors a value of $0.05$ corresponds to a multiplicative trajectory error of roughly 5\%). Crucially, sub-sampling the recorded forces to effective spacings $\Delta t$ = 50--500 leaves the logRMSE unchanged (0.045--0.052), so the residual is \textbf{not} forward-Euler discretization but a systematic property of the post-hoc reconstruction procedure.

\begin{figure}[t]\centering
  \includegraphics[width=\linewidth]{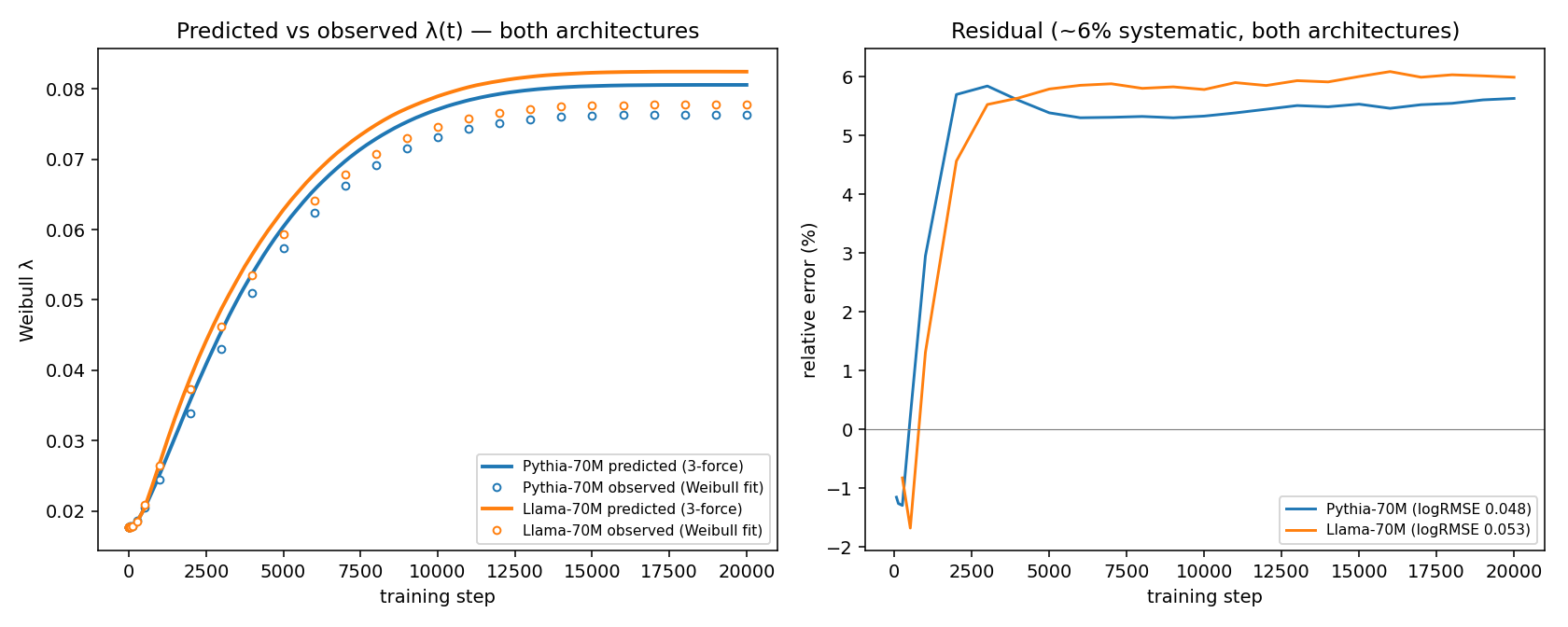}
  \caption{Closed-loop verification across architectures. Predicted $\lambda(t)$ (integrating the leading-order squared-norm recurrence) vs observed Weibull $\lambda(t)$. The observed $\lambda(t)$ is obtained by independent Weibull fits to the weight snapshots, not derived from the predicted forces, so the agreement is a genuine closed-loop check, not a self-consistency tautology. Trajectory-closure offset $\sim$4--6\%.}
  \label{fig:closure}
\end{figure}

Figure~\ref{fig:closure} provides the closed-loop verification. The predicted trajectory is obtained by integrating the measured leading-order squared-norm force recurrence and then mapping RMS to $\lambda$ through the Weibull bridge; the observed trajectory is obtained independently by fitting Weibull parameters to the saved weight checkpoints, so the agreement is not a tautology. The remaining 4--6\% trajectory offset is analyzed in Appendix~B and mainly reflects the RMS-to-Weibull bridge rather than a force-measurement failure.

Note: the logRMSE and the spline recovery rate (Pythia 0.917--0.938, Llama 0.900--0.928) are different quantities: the former measures closed-loop self-consistency; the latter measures how well the spline method recovers the alignment force from sparse checkpoints.

\subsection{Sparse Checkpoints: Spline Displacement Method}
\textbf{Problem.} Public model checkpoints expose only sparse weight snapshots $\{W(t_i)\}$, not the AdamW moments $(\hat{m}, \hat{v})$.

\textbf{Exact consecutive-step identity.} If two consecutive training states are available, the AdamW displacement identity recovers the adaptive update direction exactly:
\begin{equation}
\hat{u}_t = -(W_{t+1} - W_t)/\eta_t - \lamwd \cdot W_t
\end{equation}
The right-hand side uses only weight differences and is algebraically exact (Appendix~A).

\textbf{Sparse-checkpoint approximation.} Public checkpoints are not consecutive training states, so we estimate the missing unit-step displacement by fitting a cubic spline to the saved weight trajectory and evaluating $W(t)$ and $W(t+1)$ at unit resolution. The approximation enters only here.

\begin{figure}[t]\centering
  \includegraphics[width=0.62\linewidth]{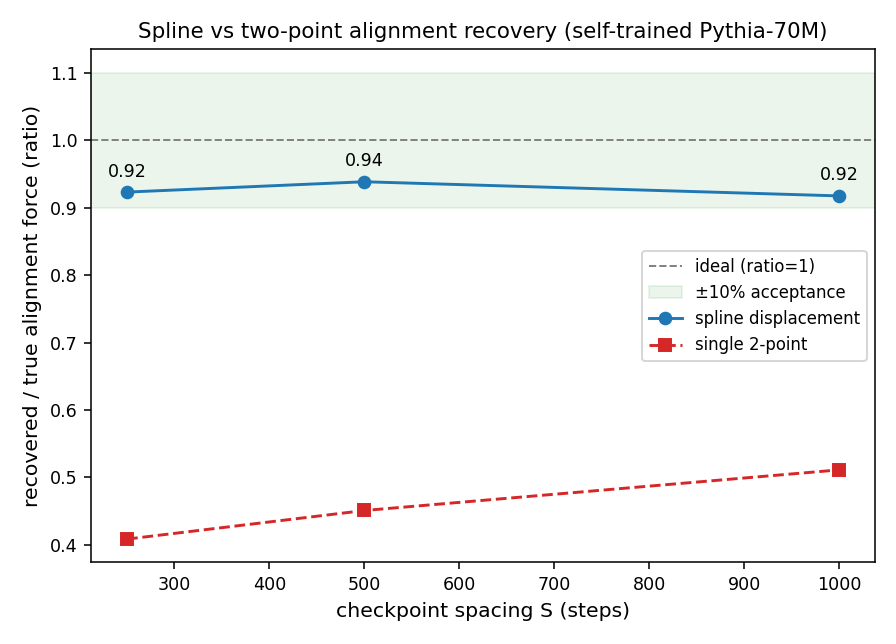}
  \caption{Spline displacement method validated on self-trained Pythia-70M (checkpoints subsampled to spacing $S$, ground-truth optimizer moments available for comparison): spline $0.917$--$0.938$ vs naive two-point baseline $0.409$--$0.511$, roughly double the recovery.}
  \label{fig:spline-verdict}
\end{figure}

\textbf{Validation.} We validate this approximation by subsampling self-trained checkpoints to intervals $S$=250/500/1000, where ground-truth optimizer moments are available for comparison (Figure~\ref{fig:spline-verdict}).

\textbf{Application.} We then apply the validated method to real Pythia checkpoints, where optimizer moments are unavailable.

\begin{figure}[t]\centering
  \includegraphics[width=0.72\linewidth]{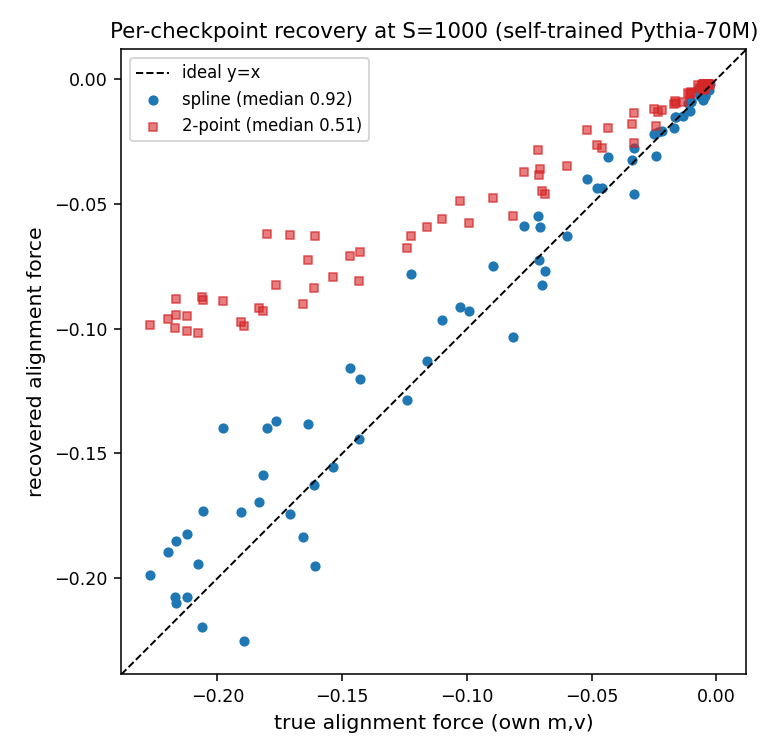}
  \caption{Per-checkpoint spline recovery (single-step force recovery, distinct from the trajectory-closure offset). Points cluster near identity with a $\sim$6--8\% systematic underestimate at high-curvature points.}
  \label{fig:spline-scatter}
\end{figure}

Figure~\ref{fig:spline-verdict} summarizes the interval-level recovery rates: across $S$=250/500/1000, spline recovery reaches 0.92--0.94 on the self-trained validation runs, roughly twice the naive two-point baseline of 0.41--0.51. Figure~\ref{fig:spline-scatter} shows the per-checkpoint recovery scatter: most points lie close to the identity line, with a systematic 6--8\% underestimate at high-curvature points. The remaining 6--8\% systematic underestimate is treated as an interpolation uncertainty bound when applying the method to real sparse checkpoints.

\section{$k$-Lock and Equilibrium}
\subsection{$k$-Lock as the Initialization Anchor}
For the present paper, the importance of the $k$-lock is methodological: it is the bridge that makes squared-norm force dynamics informative about the Weibull scale parameter $\lambda$, rather than a re-derivation of the Weibull framework itself. \citet{npmweibull2026paper1} established that $k \approx 1.20$ at initialization and remains approximately locked throughout training for all Transmission-Class components. Our per-block measurements confirm:
\begin{itemize}
  \item \textbf{Self-trained Pythia (wikitext, V1b)}: $k \in [1.178, 1.216]$ (per-block individual fit; median $\approx$ 1.200)
  \item \textbf{Self-trained Llama-style (wikitext)}: $k \in [1.192, 1.210]$ (per-block individual fit; median $\approx$ 1.203)
  \item \textbf{Real Pythia-70M (Pile, published)}: $k \in [1.148, 1.202]$ (per-block individual fit; median = \textbf{1.190})
\end{itemize}
All three are per-block individual fits, consistent with the per-block fitting protocol of \citet{npmweibull2026paper1}. Both self-trained architectures and the real published model are tightly clustered around $k_0 \approx 1.205$, with the real Pythia median of \textbf{1.190 falling in-band [1.186, 1.204]}. \textbf{The aggregate $k = 1.162$ reported across blocks is a pooling artifact; the canonical value is per-block $k \approx 1.20$.} This distinction matters because the RMS-to-$\lambda$ bridge is applied per block before aggregation; using the pooled aggregate $k$ would artificially distort the bridge.

\begin{figure}[t]\centering
  \includegraphics[width=0.82\linewidth]{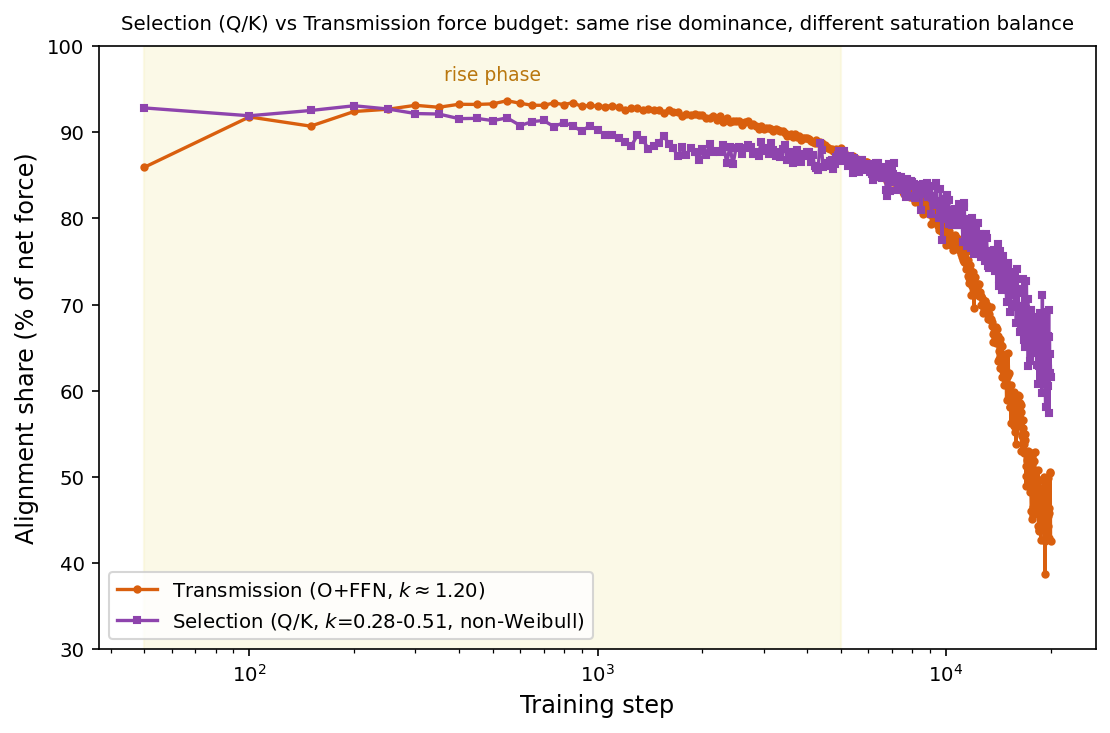}
  \caption{Three-force budget for Selection-class components (Q/K projections, Llama-style). Q and K projection matrices show different force composition from Transmission-class: both Q and K align-dominate in the rise phase, but with different magnitudes. Selection-class components do not follow Weibull$(k\approx1.20)$; $k$ ranges from 0.28 to 0.51 across layers, so force budget is reported in RMS (not $\lambda$) units. The force-balance pattern (align$\approx$decay at saturation) holds for Selection as well, suggesting the two-force transition mechanism is architecture-independent. (Saturation force shares are architecture-specific and differ from the Pythia values in Table~\ref{tab:trajectory}.)}
  \label{fig:bsel}
\end{figure}

\textbf{Selection-class check.} Figure~\ref{fig:bsel} applies the same force-budget analysis to the Q/K projection matrices in the Llama-style model. These components do not satisfy the Transmission-class $k$-lock ($k$ ranges from 0.28 to 0.51 across layers), so we report their dynamics in RMS rather than $\lambda$ units. The same qualitative alignment-to-decay balance appears at saturation, suggesting that the force-balance mechanism is a general squared-norm phenomenon, while the clean mapping to Weibull $\lambda$ is specific to $k$-locked Transmission-class components.

Since RMS and $\lambda$ are related by an approximately fixed multiplicative factor when $k$ is locked, force dynamics on $\sigma^2$ determine the dominant weight-scale component underlying the fitted Weibull $\lambda(t)$, up to the reconstruction offset quantified in Appendix~B.

\subsection{Equilibrium and the Floor}
Consistent with \citet{fan2025robust}, who observe that the terminal weight scale tracks $\sqrt{\eta/\lamwd}$, our force analysis explains this as the equilibrium condition: when $F_{\mathrm{align}} = |F_{\mathrm{decay}}|$, the net force vanishes and $\lambda$ reaches its steady-state value.

\textbf{Two-parameter structure.} The force dynamics involve two distinct regimes:
\begin{itemize}
  \item \textbf{Rise/transient phase:} dominated by the alignment force, which reflects how coherently the optimizer moves weights along their current direction --- a property of the loss landscape and training data, not a fixed parameter.
  \item \textbf{Saturation/floor phase:} the floor is maintained by a near-balance between alignment and decoupled weight decay, with decoupled weight decay ($F_{\mathrm{decay}} = -2\eta\lamwd \cdot \Sigma W^2$) providing the scale-setting counterforce. The equilibrium condition corresponds to the Fan et al. steady state, where $\lambda^* \propto \sqrt{\eta/\lamwd}$.
\end{itemize}

This equilibrium picture explains the scale-setting role of learning rate and weight decay near the floor; it does not by itself determine the transient peak, which also depends on the rise-phase alignment force and, as discussed in \S7, may vary with data coherence. Consistent with this scale-setting role, the absolute peak $\lambda$ differs between our self-trained models ($\lamwd$=0.01, $T/\tau \approx 0.20$, peak $\approx$ 0.076) and real Pythia ($T/\tau \approx 1.43$, peak $\approx$ 0.035). The decay-phase floor coefficient $c = \lambda_{\mathrm{floor}}/\sqrt{\eta_{\mathrm{floor}}/\lamwd}$ --- evaluated at the end of training, using the cosine-decayed terminal learning rate $\eta_{\mathrm{floor}}=0.1\times\eta$ and the floor (not peak) $\lambda$ --- is also regime-dependent. In a high-decay run ($\lamwd$=0.2, floor $\lambda\approx0.034$, $\eta_{\mathrm{floor}}=10^{-4}$), $c \approx 0.034/\sqrt{10^{-4}/0.2} \approx 1.51$; for real Pythia at the end of Pile (floor $\lambda\approx0.022$, $\lamwd$=0.01), $c \approx 0.022/\sqrt{10^{-4}/0.01} \approx 0.22$. This coefficient is therefore \textbf{not a universal constant}.

\section{Data-Dependent Modulation of Peak $\lambda$: An Exploratory Observation}
The force analysis in \S3--\S6 explains the dominant squared-norm component underlying $\lambda$ growth. A separate question is what determines the \emph{absolute scale} of the transient peak. We treat this question as exploratory in this paper, because isolating data effects requires controlled data-mixture experiments.

\textbf{Empirical observation.} Self-trained Pythia-70M on wikitext-103 (single-domain) reaches peak $\lambda \approx$ 0.069 (from-scratch, 118M tokens). Real Pythia-70M on the Pile (multi-domain, 22 sub-domains) reaches peak $\lambda \approx$ 0.035. Under matched optimizer hyperparameters, these differ by $\sim$2$\times$. This comparison controls the optimizer but not all data-related factors, such as token repetition, epoch count, domain entropy, and warm-start history. The absolute peak $\lambda$ is modulated by at least two distinct axes---the optimizer (learning rate, \S6.2) and data coherence (this section); we do not disentangle them here.

\textbf{Continue-training intervention.} As an intervention on the data distribution, we take a real Pythia-70M checkpoint (trained on Pile, $\lambda$ = 0.023 at initialization of this phase) and continue training on wikitext-103 with the same optimizer settings. The $\lambda$ trajectory climbs from 0.023 to \textbf{0.060} at peak --- a 2.6$\times$ increase, crossing above the original Pile peak of 0.035. This provides suggestive causal evidence that data distribution can modulate peak $\lambda$, but it does not isolate which data property is responsible.

\begin{figure}[t]\centering
  \includegraphics[width=\linewidth]{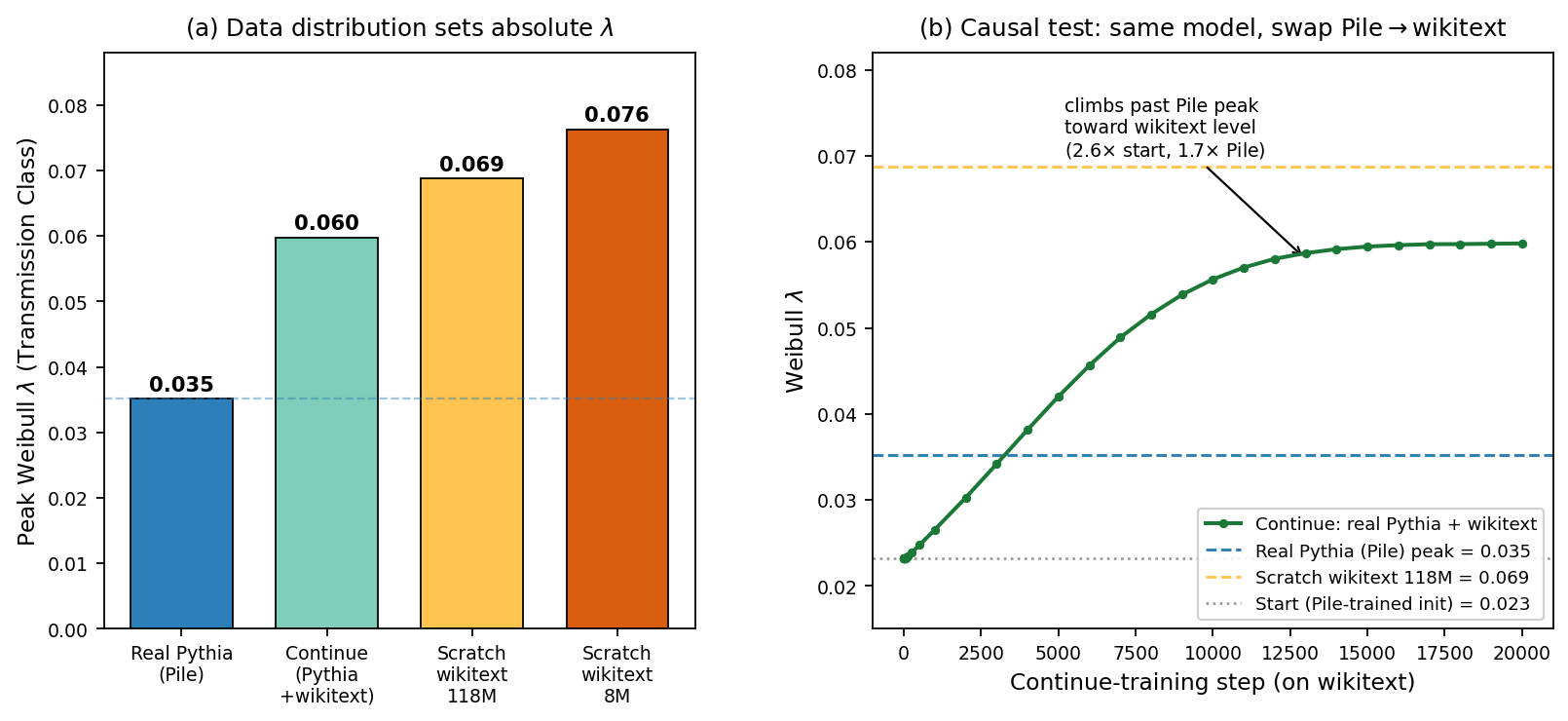}
  \caption{Training data modulates the absolute scale of $\lambda$: exploratory evidence. (a) Peak $\lambda$ across data regimes (architecture/optimizer fixed): diverse Pile $\to$ low $\lambda$ ($0.035$), single-domain wikitext $\to$ high $\lambda$ ($0.069$--$0.076$). (b) Interventional continue-training: from a Pile-trained Pythia-70M ($\lambda=0.023$), continuing on wikitext drives $\lambda$ to $0.060$ ($2.6\times$ start, $1.7\times$ Pile peak).}
  \label{fig:data-shaping}
\end{figure}

Figure~\ref{fig:data-shaping} summarizes the data-dependence of peak $\lambda$ and the interventional continue-training trajectory.

\textbf{Mechanistic interpretation.} Within the three-force framework, the alignment force is the primary driver of squared-norm growth, which in $k$-locked Transmission-class components appears as $\lambda$ growth (\S4.1). One plausible interpretation is that single-domain, highly repeated data (wikitext) induces more coherent update directions, while multi-domain data (Pile) induces partially conflicting directions; under the force-budget view, this would strengthen or weaken the alignment force and thereby change the peak weight scale. \textbf{We hypothesize that the absolute $\lambda$ scale is therefore not purely optimizer-determined; it may be modulated by the coherence structure of the training data.} This is conceptually consistent with \citet{ba2024data}, who report that increased data diversity reduces Frobenius/spectral norm in vision models --- though they analyze the eigenvalue spectrum via random matrix theory (RMT), not weight magnitudes via Weibull, and their evidence is correlational (vision augmentation), not causal (continue-training on LM pretraining).

\textbf{Quantitative summary.}

\begin{tabular}{lcc}
\toprule
Configuration & Data & peak $\lambda$ \\
\midrule
Self-trained (from scratch) & wikitext-103 (single-domain, 118M tokens) & 0.069 \\
Self-trained (from scratch) & wikitext-103 (single-domain, 8M tokens, 31 epochs) & 0.076 \\
Real Pythia (continue-train) & wikitext-103 (after Pile) & 0.060 \\
Real Pythia (published) & Pile (multi-domain) & 0.035 \\
\bottomrule
\end{tabular}

The ordering is consistent with the data-coherence hypothesis, but the table should be read as exploratory, because the rows differ in more than one data factor (repetition, epoch count, domain entropy, warm-start history). This also suggests a possible connection to the memorization literature \citep{tirumala2022memorization}: high peak $\lambda$ may reflect memorization pressure at the weight-distribution level, but direct memorization measurements are needed to test this link.

\textbf{Honest limitations.} This section identifies a \textbf{pilot observation} of data-dependent modulation of $\lambda$, but a complete characterization requires more than two corpora. \textbf{We position this as an exploratory finding}; a controlled study across 4--6 corpora spanning the single-domain$\leftrightarrow$multi-domain spectrum is reserved for a dedicated follow-up study. The continue-train evidence is interventional but starts from a warm-initialized model (Pile-trained weights) --- the trajectory reaches $\sim$87\% of the from-scratch 118M wikitext peak ($0.069$). A full account of the data-diversity axis is reserved for future work. We therefore do not treat data coherence as a primary contribution of this paper: the primary contribution remains the force-budget explanation of weight-scale dynamics, and the data result motivates a controlled follow-up study.

\section{Discussion}
\subsection{Three Alignment Objects}
Because the term ``alignment'' is overloaded in recent optimization and representation-learning literature, we explicitly distinguish the object used in this paper from two related notions:
\begin{enumerate}
  \item $\langle W, \hat{u}\rangle$ (this paper): inner product between the weight vector and the effective AdamW update direction.
  \item $\alpha_{\Delta W}$ \citep{kosson2025weight}: alignment between the weight update $\Delta W$ and the input activation $X$.
  \item NFM $\leftrightarrow$ AGOP \citep{beaglehole2024feature}: alignment between the Neural Feature Matrix and the Average Gradient Outer Product.
\end{enumerate}
These are different quantities.

\subsection{Relationship to Prior Work}
\textbf{Three-force identity.} The algebraic decomposition has been noted in prior work \citep{chou2025correction, kosson2024rotational}. We do not claim this identity as new. Our contribution is the \emph{direct measurement} of these forces on self-trained transformer models with ground-truth optimizer moments, and the checkpoint-level application to real published Pythia models through the validated spline recovery method, with the discovery that alignment dominates the rise phase.

\textbf{GSNR and injection force.} The injection force reflects gradient noise. Related work \citep{balles2018dissecting} studies the gradient signal-to-noise ratio (GSNR); consistent with this literature, we find the injection force is approximately constant ($\sim$4\%) throughout training.

\textbf{Weight scale and $\lambda_{\mathrm{wd}}$.} \citet{fan2025robust} observe that the terminal weight scale tracks $\sqrt{\eta/\lamwd}$. Our force analysis provides the mechanistic basis: it is the condition $F_{\mathrm{align}} = |F_{\mathrm{decay}}|$. Our results extend this: the floor is strongly shaped by optimizer hyperparameters, while the transient \emph{peak} may additionally reflect properties of the training data.

\textbf{Our novel contribution} is connecting leading-order AdamW squared-norm dynamics to the Weibull $\lambda(t)$ trajectory through the $k$-lock bridge, and quantifying the remaining RMS-to-Weibull reconstruction offset, explaining the overshoot$\to$relaxation pattern as a consequence of the alignment-to-decay transition.

On self-trained validation runs with ground-truth optimizer moments, the spline displacement method recovers approximately 92--94\% of the alignment force; the remaining 6--8\% systematic underestimate is treated as an interpolation uncertainty when applying the method to sparse public checkpoints.

\textbf{Limitations.} The force decomposition is measured directly only on self-trained 70M-scale models where optimizer moments are available. Real published models require sparse-checkpoint recovery, whose accuracy is validated by subsampling self-trained runs but cannot be directly verified without optimizer states. The Weibull connection also relies on the $k$-lock bridge and therefore applies most cleanly to Transmission-class components. Finally, the data-coherence observation is exploratory and requires controlled data-mixture experiments.

In this sense, the Weibull framework provides the distribution-level coordinate system, the present paper explains the AdamW force budget that moves the model along the $\lambda$ coordinate, and future controlled data-mixture studies can test how data distribution modulates that force budget.

\clearpage
\appendix
\section{An AdamW Displacement Identity and Higher-Order Terms}
We use the PyTorch-convention decoupled AdamW update:
\begin{equation}
W_{t+1} = W_t - \eta_t \cdot \hat{u}_t - \eta_t \cdot \lambda_{\mathrm{wd}} \cdot W_t \tag{A.1}
\end{equation}
where $\hat{u}_t \equiv \hat{m}_t / (\sqrt{\hat{v}_t} + \varepsilon)$.

\textbf{Displacement identity (exact).} Rearranging (A.1) for $\hat{u}_t$:
\begin{equation}
\hat{u}_t = -(W_{t+1} - W_t)/\eta_t - \lambda_{\mathrm{wd}} \cdot W_t \tag{A.2}
\end{equation}
\textbf{This is algebraically exact, not a fit or approximation.} Given two consecutive training states and $(\eta_t, \lambda_{\mathrm{wd}})$, the adaptive update direction is fully determined. Public checkpoints are usually not consecutive training states; this is why interpolation is needed in the sparse-checkpoint setting.

The squared-norm rate is:
\begin{equation}
\Delta\Sigma W^2_t = -2\eta_t\langle W_t, \hat{u}_t\rangle + \eta_t^2\|\hat{u}_t\|^2 - 2\eta_t\lambda_{\mathrm{wd}}\Sigma W^2_t + O(\eta_t^2\lambda_{\mathrm{wd}}) \tag{A.3}
\end{equation}

Substituting (A.2) into the alignment inner product:
\begin{equation}
\langle W_t, \hat{u}_t\rangle = -(1/\eta_t)\langle W_t, \Delta W_t\rangle - \lambda_{\mathrm{wd}}\|W_t\|^2 \tag{A.4}
\end{equation}

The alignment force is recoverable from weight snapshots alone:
\begin{equation}
F_{\mathrm{align},t} = 2\langle W_t, \Delta W_t\rangle + 2\eta_t\lambda_{\mathrm{wd}}\|W_t\|^2 \tag{A.5}
\end{equation}

Public checkpoints are \emph{sparse}. The spline method estimates the missing unit-step displacement by fitting a cubic spline $\widehat{W}(\cdot)$ to the saved weight trajectory and taking
\begin{equation}
\widehat{\Delta W}_t = \widehat{W}(t+1) - \widehat{W}(t) \tag{A.6}
\end{equation}
where $\widehat{W}$ is the cubic spline interpolant. This separates three distinct statements: (i) for recovering the adaptive update direction from consecutive states, the displacement identity (A.2) is exact; (ii) in sparse-checkpoint recovery, the approximation enters only through the spline estimate (A.6) of the missing unit-step displacement; and (iii) separately, the squared-norm force budget used in the main text is a leading-order decomposition that drops the higher-order AdamW decay-interaction terms quantified below.

\textbf{Scope and caveats:}
\begin{enumerate}
  \item \textbf{Identity vs. estimate.} (A.1)--(A.5) are exact algebra. Only (A.6) is an approximation. The $\sim 92\%$ recovery is a property of (A.6), not of the force decomposition.
  \item \textbf{$\varepsilon$ term.} (A.2) absorbs $\varepsilon$ into $\hat{u}$ exactly; no error is introduced.
  \item \textbf{Schedule knowledge.} (A.5) requires $(\eta_t, \lambda_{\mathrm{wd}})$.
  \item \textbf{Aggregate vs per-block.} The identity holds per-tensor. Forces in the main text are aggregated over the Transmission class ($\Sigma W^2$ over O+FFN), consistent with the Transmission-class aggregation of \citet{npmweibull2026paper1}, distinct from the per-block $k$ report (\S6.1).
    \item \textbf{Higher-order terms.} (A.3) drops $O(\eta^2\lambda_{\mathrm{wd}})$ terms, numerically negligible.

\medskip
\noindent\textbf{Higher-order term quantification.} The dropped terms are:
\begin{itemize}
  \item \textbf{Cross term:} $+2\eta_t^2\lambda_{\mathrm{wd}}\langle W_t, \hat{u}_t\rangle$
  \item \textbf{WD$^2$ term:} $+\eta_t^2\lambda_{\mathrm{wd}}^2\|W_t\|^2$
\end{itemize}
The cross term can be signed because $\langle W_t, \hat{u}_t\rangle$ can be positive or negative; the reported magnitude is negligible relative to the absolute leading-order force budget. Measured on self-trained Pythia-70M (V1b) across three training phases (all percentages reported relative to the absolute force budget):

\begin{tabular}{lcccc}
\toprule
Phase & Alignment & Cross term & WD$^2$ term & Total \\
\midrule
Early rise (step 250) & 93.1\% & $<10^{-3}$\% & $<10^{-3}$\% & 100\% \\
Mid-rise (step 2,500) & 92.1\% & $<10^{-3}$\% & $<10^{-3}$\% & 100\% \\
Saturation (step 20,000) & 49.9\% & $<10^{-3}$\% & $<10^{-3}$\% & 100\% \\
\bottomrule
\end{tabular}

Both higher-order terms contribute $<0.001\%$ of the absolute force budget across all phases (because $\eta\lambda_{\mathrm{wd}} \sim 10^{-5}$). The leading-order approximation is numerically excellent.
\end{enumerate}

\section{RMS-to-Weibull Bridge and Closed-Loop Error Bounds}
\subsection{Reconstruction Offset Decomposition}
\begin{figure}[t]\centering
  \includegraphics[width=\linewidth]{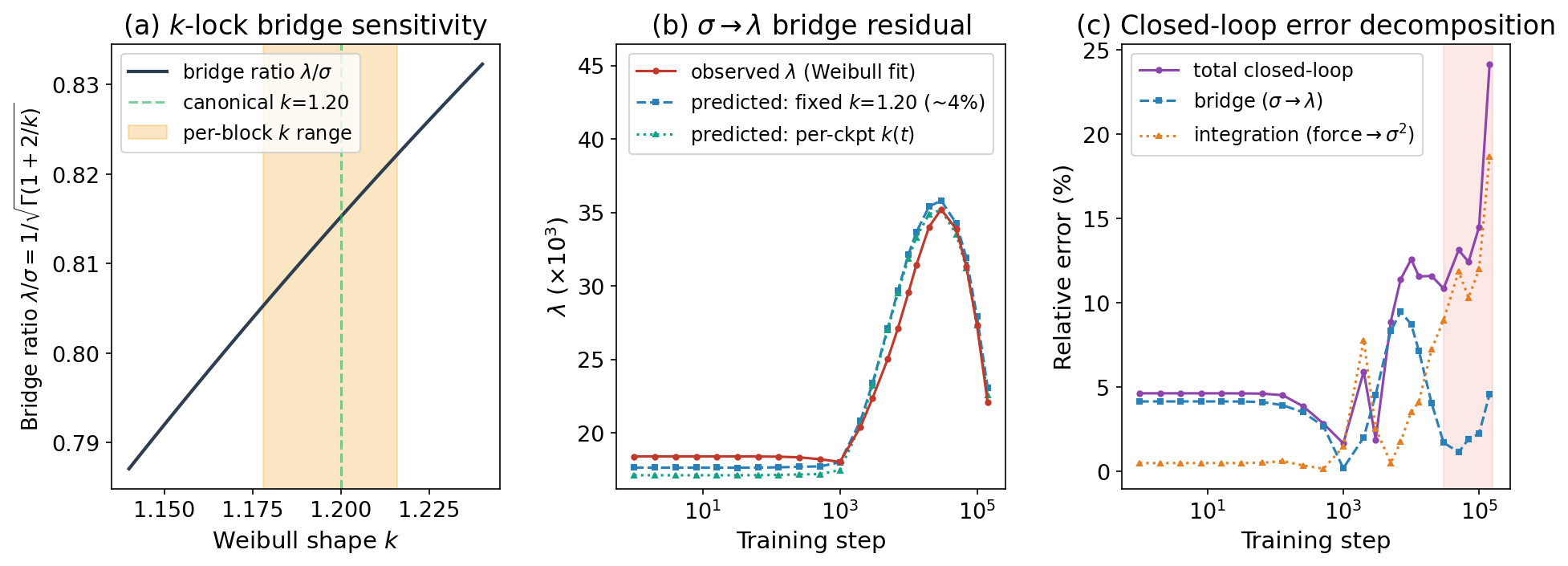}
  \caption{Bridge sensitivity and closed-loop error decomposition (real Pythia-70M). (a) The $\sigma\to\lambda$ bridge ratio $1/\sqrt{\Gamma(1+2/k)}$ vs shape $k$; over the per-block $k$ range [1.178, 1.216] (shaded, per-layer fits across all blocks) it varies $\sim$2\%, so the $k$-lock approximation is mild. (b) Observed Weibull $\lambda(t)$ vs predicted from $\sigma_{\mathrm{obs}}$ using fixed $k$=1.20 ($\sim$4\% residual) and per-checkpoint $k(t)$; the residual is dominated by the Weibull-fit nonlinearity, not $k$-variation. (c) Closed-loop relative error decomposed into bridge ($\sigma\to\lambda$) and integration (force$\to\sigma^2$) components; in the densely-sampled region (steps $\leq$20k) the total is $\sim$6\% (bridge 4.6\% + integration 1.9\%, shaded). Beyond step 30k (sparse published checkpoints, gaps 20k--43k) the pointwise error grows to 15--24\% as an integration phase-shift artifact (per-point error correlates with checkpoint gap at $r=0.86$), consistent with a sampling-induced phase-shift rather than a failure of the force-budget mechanism.}
  \label{fig:bridge}
\end{figure}

The closed-loop reconstruction integrates the leading-order squared-norm recurrence forward from the initial squared norm, maps the result to $\lambda$ through the Weibull bridge, and compares this predicted $\lambda(t)$ to the independently-fitted Weibull $\lambda(t)$. As a sanity check on the numerical integration alone, the recurrence run on a dense synthetic trajectory (V1a, $N$=1M) gives logRMSE = 0.003. On the self-trained transformers the systematic offset is $\sim$5--6\% (logRMSE $\approx$ 0.05), and on real Pythia-70M it is $\sim$8\%; for real Pythia this is a checkpoint-level closure error using spline-recovered forces, not a direct force-measurement error, because optimizer moments are unavailable. Figure~\ref{fig:bridge} summarizes the sensitivity of the $\sigma\to\lambda$ bridge to the shape $k$ and decomposes the closed-loop error into its bridge and integration components.

\textbf{Decomposition (densely-sampled region).} Over the densely-sampled region (steps $\leq$ 20k, checkpoint gaps $\leq$ 7k), the $\sim$6\% offset decomposes into:
\begin{itemize}
  \item \textbf{Bridge error} ($\sigma\to$ Weibull $\lambda$): comparing $\sigma_{\mathrm{obs}}(t)/\sqrt{\Gamma(1+2/k)}$ at fixed $k=1.20$ to fitted $\lambda_{\mathrm{obs}}(t)$ gives $\sim$4.6\%.
  \item \textbf{Integration error} (force recurrence $\to\sigma^2$): comparing integrated $\sigma_{\mathrm{pred}}(t)$ to $\sigma_{\mathrm{obs}}(t)$ gives only $\sim$1.9\%.
\end{itemize}

\textbf{Note on the Gamma factor.} For a Weibull$(k, \lambda)$ distribution, RMS and $\lambda$ are related by $\mathrm{RMS} = \lambda\sqrt{\Gamma(1 + 2/k)}$. For $k\approx1.20$, this factor is approximately 1.23, so RMS exceeds $\lambda$ by approximately 23\%; this fixed scaling is not an error. The $\sim$4.6\% bridge error is the residual between the $\lambda$ inferred from RMS using this second-moment identity and the $\lambda$ obtained by the middle-80\% Weibull probability-plot fit.

\textbf{Tail behavior (sparse checkpoints).} Beyond step 30k the published Pythia checkpoints become sparse (gaps of 20k--43k steps), and the pointwise reconstruction error grows to 15--24\%. This is a \textbf{sampling artifact}, not a mechanism failure: per-point error correlates with checkpoint gap at $r=0.86$, reflecting integration phase-shift. Denser checkpoints remove it. We report the $\sim$6\% densely-sampled offset as representative.

\subsection{Role of the $k$-Lock}
Because $k$ is not perfectly constant, the fixed-$k=1.20$ bridge incurs an additional error relative to a per-checkpoint $k(t)$ bridge. Over the per-block $k$ range $[1.178, 1.216]$ (the range of individual-layer $k$ fits), the bridge ratio varies by $\sim$2.08\%; the $k$-lock contributes $\sim$2\% to the bridge error (not $<$0.2\% as previously stated). A per-checkpoint $k(t)$ correction improves the peak fit (1.7\%$\to$0.1\%) but not the mean, confirming the residual is dominated by Weibull-fit nonlinearity, not by $k$-variation.

\textbf{Bearing on conclusions.} This reconstruction offset does not affect the force-budget conclusions, such as alignment dominance and the alignment-to-decay transition, because these are measured directly from optimizer moments in the self-trained runs rather than inferred through closed-loop integration. The $k$-lock itself is measured independently from Weibull fits to weight snapshots. In the self-trained dense setting the offset is also $\Delta t$-independent (varying $\Delta t$ from 50 to 500 leaves logRMSE unchanged at 0.045--0.052); the sparse-tail behavior of real Pythia is discussed above.

\clearpage
\section{Reproducibility Statement}
This paper builds on the NPM-Weibull diagnostic framework and its associated database \citep{npmweibull2026paper1}. The following are specific to the experiments in this paper.

\subsection{Self-Training Configurations}
\textbf{Self-trained Pythia-70M (V1b).} We train a GPT-NeoX model with the Pythia-70M configuration from random initialization on wikitext-103 (v1, train split). Optimizer: AdamW \citep{loshchilov2019decoupled} with learning rate $1\times10^{-3}$, weight decay $\lambda_{\mathrm{wd}}=0.01$, gradient decay factors $\beta_1=0.9$, $\beta_2=0.999$, warmup 200 steps followed by cosine decay to $0.1\times$ of the initial rate, batch size 24, sequence length 512, 20,000 training steps, checkpoints saved every 250 steps. Four random seeds are used.

\textbf{Self-trained Llama-style (70m).} Same hyperparameters as above, with a Llama-style architecture (SwiGLU activations, RMSNorm, full RoPE, no bias). One random seed.

\textbf{Data processing.} Tokenization uses the GPT-NeoX/Pythia tokenizer, sequence length 512, with non-overlapping contiguous chunks from the wikitext-103 train split. The 118M-token configuration trains for $\approx$2.1 epochs; the 8M-token configuration trains for 31 epochs (higher token repetition).

\subsection{Additional Controlled Runs}
\textbf{Learning-rate sweep.} We repeat the self-trained Pythia-70M setup with learning rates $\{3\times10^{-4}, 1\times10^{-3}, 3\times10^{-3}\}$, keeping model, data, weight decay, batch size, sequence length, warmup, and checkpoint frequency fixed.

\textbf{High-decay run.} For the floor-coefficient analysis, we run the same setup with $\lambda_{\mathrm{wd}}=0.2$ and report the terminal floor $\lambda$ after cosine decay.

\textbf{Selection-class force budget.} For the Llama-style model, we also compute the force budget for the Q/K projection matrices. Because these components do not satisfy the Transmission-class $k$-lock, results are reported in RMS units rather than Weibull $\lambda$ units.

\textbf{Spline subsampling validation.} To validate sparse-checkpoint recovery, we subsample self-trained runs to checkpoint intervals $S=250, 500, 1000$ and compare recovered alignment forces against ground-truth optimizer moments.

\textbf{Continue-training intervention.} We start from a published Pythia-70M checkpoint ($\lambda\approx0.023$) and continue training on wikitext-103 with the same optimizer settings as the self-trained run (learning rate $1\times10^{-3}$, $\lambda_{\mathrm{wd}}=0.01$, $\beta=(0.9,0.999)$, batch 24, sequence length 512), reinitializing the optimizer states (warm-start from weights only), using the Pythia tokenizer, for 20,000 steps with checkpoints every 250 steps.

\subsection{Real Model Checkpoints}
Real Pythia checkpoints (EleutherAI, trained on the Pile) are obtained from the EleutherAI Pythia HuggingFace repositories for the 70M, 160M, 410M, and 1B sizes (\url{https://huggingface.co/EleutherAI/pythia-{70m,160m,410m,1b}}). Published checkpoint intervals: 1,000 steps (steps $\leq$20k), 10,000 steps (steps 20k--100k), 43,000 steps (steps 100k--143k).

\subsection{Code and Data}
\begin{itemize}
  \item \textbf{Weibull fitting:} npm-weibull-py (v0.4), available at \url{https://pypi.org/project/npm-weibull-py/}, using the middle-80\% trim protocol of \citet{npmweibull2026paper1}.
  \item \textbf{Force decomposition and spline recovery:} released in the \texttt{Weibull\_WeightScale\_dynamics/} folder of the companion repository upon publication; the spline displacement method uses cubic spline interpolation via \texttt{scipy.interpolate.CubicSpline}.
  \item \textbf{NPM-Weibull database \citep{npmweibull2026paper1}:} available at \url{https://huggingface.co/datasets/TiexinDing/NPM-Weibull-DATABASE-v9_1}.
\end{itemize}

\subsection{Hardware}
Self-training runs were executed on NVIDIA GPUs (A800 80GB or RTX 4090 24GB). Total approximate compute for the self-training experiments is $\sim$40 GPU-hours.

\subsection{Random Seeds}
Four random seeds (1, 2, 3, 4) are used for the primary Pythia-70M (V1b) result; one seed for the Llama-style run. The seed is set via PyTorch \texttt{torch.manual\_seed} and NumPy \texttt{np.random.seed}.

\subsection{Software}
Experiments use Python 3, PyTorch (CUDA build), the GPT-NeoX/Llama architectures via Transformers 5.3.0, scipy (\texttt{scipy.interpolate.CubicSpline} for the spline displacement method), and npm-weibull-py v0.4 for Weibull fitting. Exact patch versions are included in the released configuration files.

\clearpage
\bibliographystyle{tmlr}
\bibliography{references}

\begin{thebibliography}{10}
\providecommand{\natexlab}[1]{#1}
\providecommand{\url}[1]{\texttt{#1}}
\expandafter\ifx\csname urlstyle\endcsname\relax
  \providecommand{\doi}[1]{doi: #1}\else
  \providecommand{\doi}{doi: \begingroup \urlstyle{rm}\Url}\fi

\bibitem[Ba et~al.(2024)Ba, Mancenido, and Pan]{ba2024data}
Yang Ba, Michelle~V. Mancenido, and Rong Pan.
\newblock Data diversity as implicit regularization: How does diversity shape
  the weight space of deep neural networks?
\newblock \emph{arXiv preprint arXiv:2410.14602}, 2024.

\bibitem[Balles \& Hennig(2018)Balles and Hennig]{balles2018dissecting}
Lukas Balles and Philipp Hennig.
\newblock Dissecting {Adam}: The sign, magnitude and variance of stochastic
  gradients.
\newblock In \emph{Proceedings of the 35th International Conference on Machine
  Learning (ICML)}, volume~80 of \emph{Proceedings of Machine Learning
  Research}, pp.\  404--413. PMLR, 2018.
\newblock arXiv:1705.07774.

\bibitem[Beaglehole et~al.(2024)Beaglehole, Mitliagkas, and
  Agarwala]{beaglehole2024feature}
Daniel Beaglehole, Ioannis Mitliagkas, and Atish Agarwala.
\newblock Feature learning as alignment: a structural property of gradient
  descent in non-linear neural networks.
\newblock \emph{Transactions on Machine Learning Research}, 2024.
\newblock ISSN 2835-8856.
\newblock URL \url{https://openreview.net/forum?id=JXCe2ZcUXr}.
\newblock arXiv:2402.05271.

\bibitem[Chou(2025)]{chou2025correction}
Jason Chuan-Chih Chou.
\newblock Correction of decoupled weight decay, 2025.
\newblock arXiv:2512.08217.

\bibitem[Ding(2026)]{npmweibull2026paper1}
Tiexin Ding.
\newblock A two-parameter {Weibull} framework for diagnosing transformer weight
  distributions.
\newblock arXiv:2605.18898 [cs.LG], 2026.
\newblock doi:10.48550/arXiv.2605.18898.

\bibitem[Fan et~al.(2025)Fan, Liu, Zhao, Yuan, and Gu]{fan2025robust}
Zhiyuan Fan, Yifeng Liu, Qingyue Zhao, Angela Yuan, and Quanquan Gu.
\newblock Robust layerwise scaling rules by proper weight decay tuning.
\newblock \emph{arXiv preprint arXiv:2510.15262}, 2025.

\bibitem[Kosson et~al.(2024)Kosson, Messmer, and Jaggi]{kosson2024rotational}
Atli Kosson, Bettina Messmer, and Martin Jaggi.
\newblock Rotational equilibrium: How weight decay balances learning across
  neural networks.
\newblock In \emph{International Conference on Machine Learning (ICML)}, pp.\
  25333--25369, 2024.
\newblock arXiv:2305.17212.

\bibitem[Kosson et~al.(2026)Kosson, Welborn, Liu, Jaggi, and
  Chen]{kosson2025weight}
Atli Kosson, Jeremy Welborn, Yang Liu, Martin Jaggi, and Xi~Chen.
\newblock Weight decay may matter more than {\textmu}{P} for learning rate
  transfer in practice.
\newblock In \emph{International Conference on Learning Representations
  (ICLR)}, 2026.
\newblock arXiv:2510.19093.

\bibitem[Loshchilov \& Hutter(2019)Loshchilov and
  Hutter]{loshchilov2019decoupled}
Ilya Loshchilov and Frank Hutter.
\newblock Decoupled weight decay regularization.
\newblock In \emph{International Conference on Learning Representations
  (ICLR)}, 2019.
\newblock arXiv:1711.05101.

\bibitem[Tirumala et~al.(2022)Tirumala, Markosyan, Zettlemoyer, and
  Aghajanyan]{tirumala2022memorization}
Kushal Tirumala, Aram~H. Markosyan, Luke Zettlemoyer, and Armen Aghajanyan.
\newblock Memorization without overfitting: Analyzing the training dynamics of
  large language models.
\newblock In \emph{Advances in Neural Information Processing Systems 35
  (NeurIPS 2022)}, pp.\  38274--38290, 2022.
\newblock arXiv:2205.10770.

\end{thebibliography}

\end{document}